\pdfoutput=1

\documentclass[11pt]{article}

\usepackage{acl}

\usepackage{times}
\usepackage{latexsym}

\usepackage[T1]{fontenc}

\usepackage[utf8]{inputenc}

\usepackage{microtype}

\usepackage{tikz-dependency}
\tikzset{/depgraph/.cd,/depgraph/.search also = {/tikz},
    baseline=-0.6ex, inner sep=-0.1cm, edge horizontal padding=3pt, edge unit distance=1.5ex}

\title{Semgrex and Ssurgeon, Searching and Manipulating Dependency Graphs}

\author{John Bauer \\
  HAI \\
  Stanford University \\
  \begin{small}\texttt{horatio@cs.stanford.edu}\end{small}
  \And
  Chlo\'e Kiddon \\
  Dept of Computer Science \\
  Stanford University \\
  \begin{small}\texttt{chloe.kiddon@gmail.com}\end{small}
  \Thanks{\ Chlo\'e Kiddon is now at Google Research.}
  \And
  Eric Yeh \\
  SRI International \\
  \begin{small}\texttt{eric.yeh@sri.com}\end{small}
  \AND
  Alex Shan \\
  Dept of Computer Science \\
  Stanford University \\
  \begin{small}\texttt{azshan@stanford.edu}\end{small}
  \And
  Christopher D. Manning \\
  Linguistics \& Computer Science \\
  Stanford University \\
  \begin{small}\texttt{manning@stanford.edu}\end{small}
}

\begin{document}
\maketitle
\begin{abstract}
Searching dependency graphs and manipulating them can be a time consuming and challenging task to get right.  We document \emph{Semgrex}, a system for searching dependency graphs, and introduce \emph{Ssurgeon}, a system for manipulating the output of \emph{Semgrex}.  The compact language used by these systems allows for easy command line or API processing of dependencies.  Additionally, integration with publicly released toolkits in Java and Python allows for searching text relations and attributes over natural text.
\end{abstract}

\section{Introduction}

With the rapid growth in languages supported by Universal
Dependencies \citep{nivre-etal-2020-universal}, being able to easily
and quickly search over dependency graphs greatly simplifies
processing of UD datasets.  Tools which allow for searching of
specific relation structures greatly simplify the work of linguists
interested in specific syntactic constructions and researchers
extracting relations as features for downstream tasks.  Furthermore,
converting existing dependency treebanks from non-UD sources to match
the UD format is valuable for adding additional data to Universal
Dependencies, and doing this conversion automatically greatly reduces
the time needed to add more datasets to UD.  Accordingly, several
tools have been developed for searching, displaying, and converting
existing datasets.

In this paper, we describe \emph{Semgrex}, a tool which searches for
regex-like patterns in dependency graphs, and \emph{Ssurgeon}, a tool to
rewrite dependency graphs using the output of \emph{Semgrex}.  Both
systems are written in Java, with Java API and command line tools
available.  In addition, there is a Python interface, including using
displaCy \citep{Honnibal_spaCy_Industrial-strength_Natural_2020} as a
library to visualize the results of searches and transformation
operations.  The tools can be used programmatically to enable further
processing of the results or used via the included command line tools.
Furthermore, a web interface\footnote{\url{https://corenlp.run/}} shows the
results of applying patterns to raw text.

\emph{Semgrex} was released as part of CoreNLP
\citep{manning-etal-2014-stanford}.  As such, it has been
used in several projects (section \ref{sec:usage}).  Several existing
uses of \emph{Semgrex} make use of it via the API, one of the strengths of
the system, such as the OpenIE relation extraction
of \citep{angeli-etal-2015-leveraging}.

More recently,
we have added new pattern matching capabilities such as exact edge
matching, a new python interface, and performance improvements.  The
previously unpublished \emph{Ssurgeon} adds useful new capabilities
for transforming dependency graphs.

Many of the features described here are similar to those in
Grew-Match \citep{guillaume-2021-graph} and
Semgrex-Plus \citep{tamburini-2017-semgrex}.  Similar to \emph{Ssurgeon},
Semgrex-Plus uses \emph{Semgrex} to find matches for editing, but \emph{Ssurgeon}
supports a wider range of operations.  Compared with Grew-Match,
\emph{Semgrex} and \emph{Ssurgeon} have the ability to start with raw text and
search for dependency relations directly.

\section{Semgrex}

\emph{Semgrex} and \emph{Ssurgeon} are publicly released as part of the CoreNLP software package \citep{manning-etal-2014-stanford}\footnote{\url{https://stanfordnlp.github.io/CoreNLP/}}.  \emph{Semgrex} reads dependency trees from CoNLL-U files or parses dependencies from raw text using the associated CoreNLP parser.

Search patterns are composed of two pieces: node descriptions and
relations between nodes.  A search is executed by iterating over
nodes, comparing each word to the node search pattern and checking its
relationships with its neighbors using the relation search patterns.

\subsection{Node Patterns}

Dependency graphs of words and their dependency relations are
represented internally using a directed graph, with nodes representing
the words and labeled edges representing the dependencies.

When searching over nodes, the most generic node description is
empty curly brackets.  This search matches every node in the graph:

\begin{verbatim}
  {}
\end{verbatim}

Within the brackets, any attributes of the words available to \emph{Semgrex} can be queried.  For example, to query a person's name:

\begin{verbatim}
  {word:Beckett}
\end{verbatim}

It is possible to use standard string regular expressions, using the match semantics, within the attributes:

\begin{verbatim}
  {word:/Jen.*/}
\end{verbatim}

A node description can be negated.  For example, this will match any
word, as long as it does not start with Jen, Jenny, Jennifer, etc.

\begin{verbatim}
  !{word:/Jen.*/}
\end{verbatim}

Node descriptions can also be combined:

\begin{verbatim}
  {word:/Jen.*/;tag:NNP}
\end{verbatim}

See Table~\ref{tab:attributes} for a list of commonly used word
attributes.  Note that \verb|ner| may require a tool which provides
NER annotations, such as \citep{manning-etal-2014-stanford} (Java) or \citep{qi2020stanza} (Python),
as those are typically not part of UD treebanks.

\begin{table}
\centering
\begin{tabular}{ccc}
\hline
\multicolumn{3}{|c|}{\textbf{Attribute}} \\
\hline
word & pos & lemma \\
ner & idx & upos \\
\end{tabular}
\caption{Commonly used attributes of words}
\label{tab:attributes}
\end{table}

\subsection{Relation Patterns}

\begin{table}
\begin{small}
\centering
\begin{tabular}{cc}
\hline
\textbf{Relation} & \textbf{Meaning} \\
\hline
A < B    & A is the dependent of B \\
A > B    & A is the governor of B \\
A <{}< B   & A is part of a chain of deps to B \\
A >{}> B   & A is part of a chain of govs to B \\
A . B    & idx(A) == idx(B) $-$ 1 \\
A .. B   & idx(A) < idx(B) \\
A - B    & idx(A) == idx(B) + 1 \\
A -{}- B   & idx(A) > idx(B) \\
A \$+ B  &  C, A < C, B < C, idx(A) == idx(B) $-$ 1 \\
A \$- B  &  C, A < C, B < C, idx(A) == idx(B) + 1 \\
A \$++ B &  C, A < C, B < C, idx(A) < idx(B) \\
A \$-{}- B &  C, A < C, B < C, idx(A) > idx(B) \\
A >++ B  & A gov B, idx(A) < idx(B) \\
A >-{}- B  & A gov B, idx(A) > idx(B) \\
A <++ B  & A dep B, idx(A) < idx(B) \\
A <-{}- B  & A dep B, idx(A) > idx(B) \\
\end{tabular}
\caption{Relations between words}
\label{tab:relations}
\end{small}
\end{table}

Adding relations between nodes allows for searching over graph
structures, making \emph{Semgrex} a powerful search tool over
dependency graphs.  The relations used can be from any dependency
formalism, although CoreNLP and Stanza both use Universal Dependencies
by default.

The simplest relations are parent and child:

\begin{verbatim}
  {word:Dep} < {word:Gov}
  {word:Gov} > {word:Dep}
\end{verbatim}

Many relations consider word order as well, such as the sister
relations: $+$ indicates the word on the left of the relation comes
first, and $-$ indicates the word on the right comes first:

\begin{verbatim}
  {word:A} $+ {word:B}
  {word:A} $- {word:B}
\end{verbatim}

\noindent See Table~\ref{tab:relations} for a list of supported relations.

Relations can have labels, in which case the types on the edge between nodes must match:
\nopagebreak
\begin{verbatim}
  {} <nsubj {}
\end{verbatim}
\nopagebreak
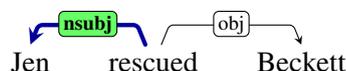
\begin{center}
\begin{dependency}
   \begin{deptext}[column sep=0.6cm]
      Jen \& rescued \& Beckett \\
   \end{deptext}
   \depedge[edge style={blue!60!black,ultra thick},
            label style={fill=green!60,font=\bfseries,text=black}]{2}{1}{nsubj}
   \depedge{2}{3}{obj}
\end{dependency}
\end{center}

A special token matches exactly at root, such as in this example from the UD conversion of EWT \citep{silveira-etal-2014-gold}:

\begin{verbatim}
  {$} > {}
\end{verbatim}
\nopagebreak
\begin{center}
\begin{dependency}
   \begin{deptext}[column sep=0.3cm]
      guerrillas \& kidnapped \& a \& cosmetic \& surgeon \\
   \end{deptext}
   \depedge[edge style={blue!60!black,ultra thick},
            label style={fill=green!60,font=\bfseries,text=black}]{2}{1}{nsubj}
   \depedge[edge style={blue!60!black,ultra thick},
            label style={fill=green!60,font=\bfseries,text=black}]{2}{5}{obj}
   \depedge{5}{3}{det}
   \depedge{5}{4}{amod}
\end{dependency}
\end{center}

Relations can be chained.  Without parentheses, subsequent relations
all apply to the same node; brackets denote that the later relations
apply between the nodes in brackets as opposed to the head of the
expression.

\begin{verbatim}
  {} >nsubj {} >obj {}
\end{verbatim}
\nopagebreak
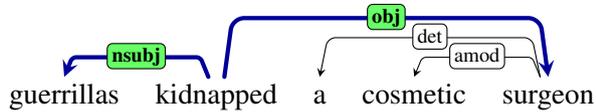
\begin{center}
\begin{dependency}
   \begin{deptext}[column sep=0.3cm]
      guerrillas \& kidnapped \& a \& cosmetic \& surgeon \\
   \end{deptext}
   \depedge[edge style={blue!60!black,ultra thick},
            label style={fill=green!60,font=\bfseries,text=black}]{2}{1}{nsubj}
   \depedge[edge style={blue!60!black,ultra thick},
            label style={fill=green!60,font=\bfseries,text=black}]{2}{5}{obj}
   \depedge{5}{3}{det}
   \depedge{5}{4}{amod}
\end{dependency}
\end{center}

\begin{verbatim}
  {} >obj ({} >amod {})
\end{verbatim}
\nopagebreak
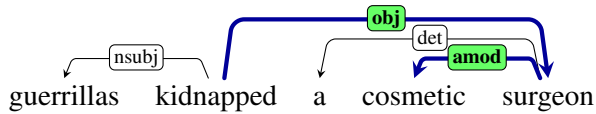
\begin{center}
\begin{dependency}
   \begin{deptext}[column sep=0.3cm]
      guerrillas \& kidnapped \& a \& cosmetic \& surgeon \\
   \end{deptext}
   \depedge{2}{1}{nsubj}
   \depedge[edge style={blue!60!black,ultra thick},
            label style={fill=green!60,font=\bfseries,text=black}]{2}{5}{obj}
   \depedge{5}{3}{det}
   \depedge[edge style={blue!60!black,ultra thick},
            label style={fill=green!60,font=\bfseries,text=black}]{5}{4}{amod}
\end{dependency}
\end{center}

More advanced conjunction and disjunction operations are also possible.  The JavaDoc\footnote{\url{https://nlp.stanford.edu/nlp/javadoc/javanlp/edu/stanford/nlp/semgraph/semgrex/SemgrexPattern.html}} reference describes the complete syntax.

\subsection{Named Nodes and Relations}

It is possible to name one or more nodes as part of a \emph{Semgrex} pattern.
This allows for relations between three or more nodes using
backreferences.  When a node is named in a backreference, it must be
the exact same node as the first instance for the pattern to match.

\begin{verbatim}
  {word:running} 
    >nsubj ({} >conj {}=C)
    >nsubj {}=C
\end{verbatim}
\nopagebreak
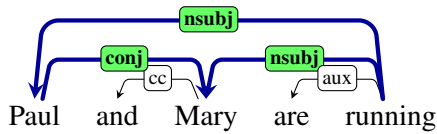
\begin{center}
\begin{dependency}
   \begin{deptext}[column sep=0.3cm]
      Paul \& and \& Mary \& are \& running \\
   \end{deptext}
   \depedge{3}{2}{cc}
   \depedge[edge style={blue!60!black,ultra thick},
            label style={fill=green!60,font=\bfseries,text=black}]{5}{3}{nsubj}
   \depedge{5}{4}{aux}
   \depedge[edge style={blue!60!black,ultra thick},
            label style={fill=green!60,font=\bfseries,text=black}]{5}{1}{nsubj}
   \depedge[edge style={blue!60!black,ultra thick},
            label style={fill=green!60,font=\bfseries,text=black}]{1}{3}{conj}
\end{dependency}
\end{center}

It is also possible to name the edges.  This is useful when combined
with \emph{Ssurgeon}, which can manipulate an edge based on its name.

\begin{verbatim}
  {word:running} 
    >nsubj ({} >conj=conj {}=C)
    >nsubj {}=C
\end{verbatim}
\nopagebreak
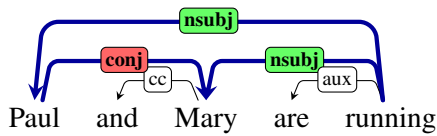
\begin{center}
\begin{dependency}
   \begin{deptext}[column sep=0.3cm]
      Paul \& and \& Mary \& are \& running \\
   \end{deptext}
   \depedge{3}{2}{cc}
   \depedge[edge style={blue!60!black,ultra thick},
            label style={fill=green!60,font=\bfseries,text=black}]{5}{3}{nsubj}
   \depedge{5}{4}{aux}
   \depedge[edge style={blue!60!black,ultra thick},
            label style={fill=green!60,font=\bfseries,text=black}]{5}{1}{nsubj}
   \depedge[edge style={blue!60!black,ultra thick},
            label style={fill=red!60,font=\bfseries,text=black}]{1}{3}{conj}
\end{dependency}
\end{center}

\subsection{Concrete Example}
\label{subsec:concrete}

Here are a couple examples from a slot filling task using \emph{Semgrex}.
Both examples search for ``son'' or ``daughter'' in relation to possible
family members.  \citep{Angeli2014StanfordsDS}

This matches ``John's daughter, Logan, \ldots''

\begin{verbatim}
{lemma:/son|daughter|child/}
 >/nmod:poss/ {ner:PERSON}=ent
 >appos {ner:PERSON}=slot
\end{verbatim}

This matches ``Tommy, son of John, \ldots''

\begin{verbatim}
{ner:PERSON}=slot
 >appos
   ({lemma:/son|daughter|child/}
    >/nmod:of/ {ner:PERSON}=ent)
\end{verbatim}

\subsection{Implementation Notes}

Under the hood, the tool is built using JavaCC\footnote{\raggedright\url{https://javacc.github.io/javacc/documentation/grammar.html}} to process input patterns.

The graphs are implemented as a collection of edges as relations, with nodes storing indices and the text information such as word, lemma, and POS.  To represent edges to hidden copy nodes, nodes can be pointers to the same underlying data with a copy index on them.  An example where this happens is \emph{I went over the river and through the woods}, where the unstated \emph{went} before \emph{through the woods} is represented as a copy node.

Nodes are searched in topographical order if possible, and in index order if not, with the intention of making a canonical ordering on the search results.

\section{Ssurgeon}

\emph{Ssurgeon} is an extension of \emph{Semgrex} which includes rules to rewrite dependency graphs.

A pattern in \emph{Ssurgeon} is a pattern for \emph{Semgrex} with required
named nodes and/or edges, depending on the edit type, along with a
edit definition.

\subsection{Edge Editing}

To add a new edge, the edit pattern must specify the governor, the
dependent, and the edge type.  The \emph{Ssurgeon} pattern will add an
edge to each match of the \emph{Semgrex} pattern.

For example, in the previous ``Paul and Mary are running'' graph, the
following would add the second $nsubj$ if it did not already exist.
This would be useful for making enhanced dependencies, as basic UD has
$conj(Paul, Mary)$ but not $nsubj(running, Mary)$.)
If the edge already exists, this rule does not add a duplicate edge.

\begin{verbatim}
  {word:running}=A
    >nsubj 
    ({}=B >conj {}=C)

  addEdge -gov A -dep C
          -reln nsubj
\end{verbatim}
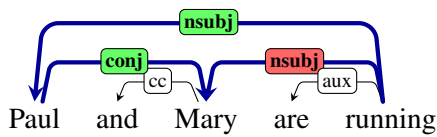
\nopagebreak
\begin{center}
\begin{dependency}
   \begin{deptext}[column sep=0.3cm]
      Paul \& and \& Mary \& are \& running \\
   \end{deptext}
   \depedge{3}{2}{cc}
   \depedge[edge style={blue!60!black,ultra thick},
            label style={fill=red!60,font=\bfseries,text=black}]{5}{3}{nsubj}
   \depedge{5}{4}{aux}
   \depedge[edge style={blue!60!black,ultra thick},
            label style={fill=green!60,font=\bfseries,text=black}]{5}{1}{nsubj}
   \depedge[edge style={blue!60!black,ultra thick},
            label style={fill=green!60,font=\bfseries,text=black}]{1}{3}{conj}
\end{dependency}
\end{center}

There are two mechanisms for deleting an edge.  The first deletes an edge between two named nodes, and the second deletes a named edge.  All edges which match the \emph{Semgrex} pattern will be deleted.

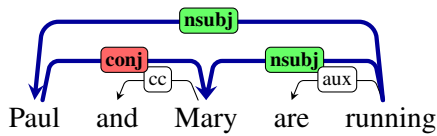
\begin{center}
\begin{dependency}
   \begin{deptext}[column sep=0.3cm]
      Paul \& and \& Mary \& are \& running \\
   \end{deptext}
   \depedge{3}{2}{cc}
   \depedge[edge style={blue!60!black,ultra thick},
            label style={fill=green!60,font=\bfseries,text=black}]{5}{3}{nsubj}
   \depedge{5}{4}{aux}
   \depedge[edge style={blue!60!black,ultra thick},
            label style={fill=green!60,font=\bfseries,text=black}]{5}{1}{nsubj}
   \depedge[edge style={blue!60!black,ultra thick},
            label style={fill=red!60,font=\bfseries,text=black}]{1}{3}{conj}
\end{dependency}
\end{center}
\nopagebreak
\begin{verbatim}
  {word:running} 
    >nsubj {}=B
    >nsubj ({}=C !== {}=B)

  removeEdge -gov B -dep C
             -reln conj
\end{verbatim}
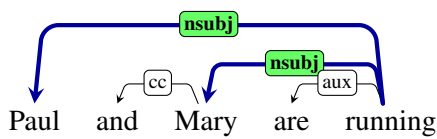
\nopagebreak
\begin{center}
\begin{dependency}
   \begin{deptext}[column sep=0.3cm]
      Paul \& and \& Mary \& are \& running \\
   \end{deptext}
   \depedge{3}{2}{cc}
   \depedge[edge style={blue!60!black,ultra thick},
            label style={fill=green!60,font=\bfseries,text=black}]{5}{3}{nsubj}
   \depedge{5}{4}{aux}
   \depedge[edge style={blue!60!black,ultra thick},
            label style={fill=green!60,font=\bfseries,text=black}]{5}{1}{nsubj}
\end{dependency}
\end{center}

Alternatively, $removeNamedEdge$ removes a labeled edge:

\begin{verbatim}
  {word:running}
    >nsubj {}=B
    >nsubj ({}=C >conj=conj {}=B)

  removeNamedEdge -edge conj
\end{verbatim}

There is also a mechanism for relabeling an edge, such as might be handy when mapping dependency graphs from one formalism to another:

\begin{verbatim}
  {word:running}
    >nsubj {}=B
    >nsubj ({}=C >conj=conj {}=B)

  relabelNamedEdge
    -edge conj -reln dep
\end{verbatim}

\subsection{Node Editing}

There are mechanisms to add a node, remove a subgraph starting from a node, and alter the content of a node.

To add a node, specify a position, a node to attach it to, and a
relation.  The position can be at the start or end of a sentence or relative
to an existing node.  For this modification, it is necessary to add a
guard to the \emph{Semgrex} expression, or it will enter an infinite loop of
adding unlimited new nodes to the graph.

Note that in the following example, searching for a word with no $det$
and adding a $det$ to the node is already sufficient to prevent runaway nodes.

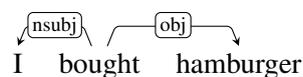
\begin{center}
\begin{dependency}
   \begin{deptext}[column sep=0.3cm]
      I \& bought \& hamburger \\
   \end{deptext}
   \depedge{2}{1}{nsubj}
   \depedge{2}{3}{obj}
\end{dependency}
\end{center}

\begin{verbatim}
  {word:bought}
    >dobj ({}=A !>det {})

  addNode
    -word=a -reln det
    -gov A -position +A
\end{verbatim}

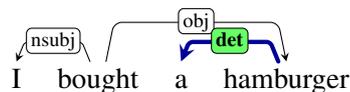
\begin{center}
\begin{dependency}
   \begin{deptext}[column sep=0.3cm]
      I \& bought \& a \& hamburger \\
   \end{deptext}
   \depedge{2}{1}{nsubj}
   \depedge{2}{4}{obj}
   \depedge[edge style={blue!60!black,ultra thick},
            label style={fill=green!60,font=\bfseries,text=black}]{4}{3}{det}
\end{dependency}
\end{center}

\section{Usage}
\label{sec:usage}

\emph{Semgrex} has been used for task-specific processing of academic work \citep{Shah2018ARA}, news summarization \citep{li-etal-2016-abstractive}, text-to-scene generations \citep{Chang2014LearningSK}, relation extraction \citep{Chaganty2017StanfordAT}, and processing medical documents \citep{Profitlich2021ACS}.

\emph{Ssurgeon} was used internally to simplify sentences using their dependencies as an extension to the textual entailment system in \citep{chambers-etal-2007-learning}.

The Java API and command line interfaces are part of the Java package CoreNLP \citep{manning-etal-2014-stanford}\footnote{\url{https://stanfordnlp.github.io/CoreNLP/}}.  The python client is available via Stanza \citep{qi2020stanza}\footnote{\url{https://stanfordnlp.github.io/stanza/}}
The code is actively maintained as of 2023, and suggestions for additional \emph{Semgrex} relations, \emph{Ssurgeon} operations, or other improvements are welcome at our github repo\footnote{\url{https://github.com/stanfordnlp/CoreNLP}}.

\subsection{Python Integration}
\label{sec:python}

Both \emph{Semgrex} and \emph{Ssurgeon} have Python APIs.  This allows for
operations on the results of Stanza \citep{qi2020stanza} on natural language or using
the API in Stanza to read and process existing UD datasets.

Here is an example of using \emph{Semgrex} on the results of parsing an
article on disease transmission with Stanza to find out what
insect vectors transmit a disease.  The search patterns here are
abridged for readability.

\begin{tiny}
\begin{verbatim}
EXPR = """
{word:/transmitted/} >/obl|advcl/
 ({word:/^(?!bite|biting|bites).*$/}=vector
  >/case|mark/ {word:/by|from|through/})
"""

def process_text(parser, text):
   doc = parser(text)
   results = semgrex.process_doc(doc, EXPR)
   facts = OrderedDict()
   for sentence_results, sentence in zip(results.result, 
                                         doc.sentences):
      if sentence.text in facts:
         # already seen this exact sentence!
         # results will be exactly the same
         continue
      facts[sentence.text] = []
      for pattern_result in results.result:
         if len(pattern_result.match) == 0:
            continue
         for match in pattern_result.match:
            for named_node in match.node:
               new_fact = "  {}: {}".format(named_node.name, 
                   sentence.tokens[named_node.index-1].text)
               if new_fact not in facts[sentence.text]:
                  facts[sentence.text].append(new_fact)
   return facts
\end{verbatim}
\end{tiny}

Also included is a mechanism to display graphs as search results,
thanks to an API call to displaCy \citep{Honnibal_spaCy_Industrial-strength_Natural_2020}.

\section{Related Work}


The analysis of dependency treebanks, especially Universal
Dependencies, has a long history of using dependency searching and
rewriting tools.


Constituency treebanks such as the Penn Treebank \citep{marcus-etal-1993-building}
predate Universal Dependencies.  To analyze such constituency datasets,
Tgrep \citep{tgrep} and its successor Tgrep2 \citep{tgrep2} set the
initial standard for searching tree structured data.  Tregex and
Tsurgeon \citep{levy-andrew-2006-tregex} extended the language and
added functionality to rewrite constituency trees.

\emph{Semgrex} was one of the earliest tools to address the problem
of searching in dependency graphs.  It was previous briefly described
in a paper on entailment that was the first to use \emph{Semgrex},
\citep{chambers-etal-2007-learning}, although that paper did not
include \emph{Ssurgeon} or fully explain the usage of \emph{Semgrex}.
\emph{Semgrex} has been used by other research several times in the following years.


The authors of the UD\_Italian-VIT
\citep{Alfieri2016AlmostAC} dataset used an extension of \emph{Semgrex},
Semgrex-Plus \citep{tamburini-2017-semgrex}, to convert the dependency
form of VIT to Universal Dependencies.  Semgrex-Plus adds
edge creation, edge deletion, and word relabeling to a \emph{Semgrex} result.


Also connected with UniversalDependencies are multiple search engines
which allow for easier viewing of the treebanks.
Tündra \citep{martens-tundra} allows for searching of a variety of
treebanks using the TIGERSearch
format \citep{TIGERsearch}.\footnote{\raggedright\url{https://www.ims.uni-stuttgart.de/documents/ressourcen/werkzeuge/tigersearch/doc/html/QueryLanguage.html}}
Additional treebanks not part of UD are
included \citep{martens-passarotti-2014-thomas}, although recent UD
updates have not been incorporated.


Grew-Match \citep{guillaume-2021-graph}\footnote{\url{http://universal.grew.fr/}}
hosts a website which allows for searching of existing UD and other dependency datasets.
The interface is frequently updated, hosting the latest 2.11 treebanks
as of January 2023.


\citep{heinecke-2019-conllueditor} provides a web interface to
backend parsers such as \citep{straka-etal-2016-udpipe} and provides
search, visual editing, and automatic pattern matching and
replacement.  However, it does not include a command line tool.


Other tools include UDeasy \citep{brigadavilla:2022:CMLC10}, which provides a graphical interface
which allows for searching of UD treebanks or any other dependency
formalism.  spaCy reimplemented \emph{Semgrex} as part of the 3.0 release, adding the
DependencyMatcher tool
\citep{Honnibal_spaCy_Industrial-strength_Natural_2020}.\footnote{\raggedright\url{https://spacy.io/api/dependencymatcher}}
UDapi \citep{Popel2017UdapiUA} provides mechanisms for searching
dependency graphs, parsing text, visualizing the graphs, and
manipulating the graphs themselves.
Odin is a rule-based event extraction framework over dependency structures \citep{ValenzuelaEscarcega2016OdinsRA}.



\section{Conclusion}

We have introduced \emph{Semgrex} and \emph{Ssurgeon}, flexible, simple systems for dependency
matching and dependency tree manipulation.

\section{Acknowledgements}

We would like to thank the reviewers for their helpful feedback on this work.

\bibliography{anthology,custom}

\end{document}